\documentclass[10pt,twocolumn,letterpaper]{article}

\usepackage[pagenumbers]{iccv} 

\definecolor{iccvblue}{rgb}{0.21,0.49,0.74}
\usepackage[pagebackref,breaklinks,colorlinks,allcolors=iccvblue]{hyperref}

\def\paperID{2} 
\def\confName{ICCV}
\def\confYear{2025}

\title{Automated Detection of Antarctic Benthic Organisms in High-Resolution In Situ Imagery to Aid Biodiversity Monitoring}

\usepackage{gensymb} 
\usepackage{graphicx} 
\usepackage{multirow} 
\usepackage{diagbox} 
\usepackage{pifont}
\newcommand{\cmark}{\ding{51}}
\newcommand{\xmark}{\ding{55}}
\usepackage{balance} 
\usepackage{float} 
\usepackage{pdflscape} 
\usepackage{subcaption} 
\usepackage{hyperref} 
\usepackage[accsupp]{axessibility}  

\author{
Cameron Trotter$^{\dagger*}$ \quad
Huw Griffiths$^{\dagger}$ \quad
Tasnuva Ming Khan$^{\dagger\ddagger}$ \quad
Rowan Whittle$^{\dagger}$\\
$^{\dagger}$British Antarctic Survey\quad $^{\ddagger}$University of Cambridge\\
\small{\texttt{cater@bas.ac.uk\thanks{Corresponding author} \quad hjg@bas.ac.uk \quad tfmk2@cam.ac.uk \quad roit@bas.ac.uk}}
}

\begin{document}
\maketitle
\begin{abstract}
Monitoring benthic biodiversity in Antarctica is vital for understanding ecological change in response to climate-driven pressures. This work is typically performed using high-resolution imagery captured in situ, though manual annotation of such data remains laborious and specialised, impeding large-scale analysis. We present a tailored object detection framework for identifying and classifying Antarctic benthic organisms in high-resolution towed camera imagery, alongside the first public computer vision dataset for benthic biodiversity monitoring in the Weddell Sea. Our approach addresses key challenges associated with marine ecological imagery, including limited annotated data, variable object sizes, and complex seafloor structure. The proposed framework combines resolution-preserving patching, spatial data augmentation, fine-tuning, and postprocessing via Slicing Aided Hyper Inference. We benchmark multiple object detection architectures and demonstrate strong performance in detecting medium and large organisms across 25 fine-grained morphotypes, significantly more than other works in this area. Detection of small and rare taxa remains a challenge, reflecting limitations in current detection architectures. Our framework provides a scalable foundation for future machine-assisted in situ benthic biodiversity monitoring research.
\end{abstract}
\vspace{-1em}    
\section{Introduction}
\label{sec:intro}

Benthic communities, comprised of organisms that live in, on or around the seafloor, are highly biodiverse, play key roles within global nutrient cycling, and are a valuable food source \citep{lam-gordilloEcosystemFunctioningFunctional2020}. Global anthropogenic change, e.g., ocean warming and acidification \citep{reddinGlobalWarmingGenerates2022, shiImpactsOceanAcidification2024}, coupled with direct local and regional pressures such as harvesting and pollution, are negatively impacting the structure and function of benthic communities \citep{crespoEcologicalEconomicImportance2022}.

The Antarctic benthos is uniquely adapted to its isolated and frozen environment \citep{alcarazBiogeographicAtlasSouthern2014}. These cold-adapted species face additional pressures through changes to the cryosphere that dominates their ocean, e.g., glacial melt and ice shelf collapse. These changes are most notable in the shallow benthic communities of the West Antarctic Peninsula, where changes to biodiversity, trophic structure, biomass, and distribution have been observed \citep{griffithsAntarcticBenthicEcological2024}. 

Historically, the exploration and monitoring of benthic environments has relied on invasive, non-quantitative methods such as dredging, or more quantitative yet slow-to-deploy instruments like corers and grabs \citep{reesGuidelinesStudyEpibenthos2009}. In recent years, the adoption of imaging technologies, delivered via SCUBA, submersibles, towed or drop camera systems, remotely operated vehicles, and autonomous platforms, has significantly increased both the rate and scale of data acquisition. Photographic and video data enable rapid, in situ, and quantitative surveys of extensive seafloor areas.

Imaging techniques represent a non-destructive and repeatable survey method to monitor ecosystem change. To date, the usefulness of collected data has been restricted by the need for expert assessment of every image, which is time consuming \citep{williamsLeveragingAutomatedImage2019, aliciaImageAnalysisBenthic2023} and prone to fatigue and annotation bias \citep{culverhouseHumanMachineFactors2007, durdenComparisonImageAnnotation2016, piechaudAutomatedIdentificationBenthic2019}. This bottleneck is particularly evident for Antarctica, with highly diverse and endemic benthic species \citep{alcarazBiogeographicAtlasSouthern2014} and relatively few taxonomic experts capable of providing confident image-based identifications.

Additionally, the high logistical and financial costs associated with deep-sea data collection often results in comparatively small amounts of collected data. Antarctica's geographic isolation and extreme environmental conditions make fieldwork highly resource-intensive, limiting collection to infrequent, short-duration missions typically led by national research programmes.

Recent advances in deep learning and computer vision have enabled the development of machine-assisted in situ biodiversity monitoring tools, designed to automate parts of the data curation process and mitigate the annotation bottleneck faced by marine ecologists \citep{trotterSurveyingDeepReview2025}. By leveraging manually curated data from previous surveys, researchers can now train models to detect benthic organisms commonly encountered in their study regions, supporting applications such as first-pass annotation workflows \citep{langenkamperBIIGLEBrowsingAnnotating2017, pedersenDetectionMarineAnimals2019, pavoniTagLabAIassistedAnnotation2022}.

Given the high data collection and annotation costs associated with the Antarctic benthos however, there is a notable lack of publicly available datasets suitable for training automated biodiversity monitoring tools for these ecosystems. This limitation is further compounded by the region’s high levels of endemism, which reduces the relevance and transferability of models trained on data from other regions.

In scenarios requiring the detection of small or densely aggregated organisms, high-resolution imagery is often utilised to enable finer-scale ecological observations. However, such data introduces additional complexity to the model development pipeline. High-resolution images place substantial computational demands on both training and inference processes, and the accurate detection of small or closely packed objects remains a persistent challenge for deep learning-based object detection systems \citep{LIU2021114602}.

In this study, we present an object detection framework designed to identify benthic organisms in high-resolution seafloor imagery from the Weddell Sea, Antarctica. Our approach accommodates large-scale inputs without downscaling through a patch-based processing methodology. The model is trained on a manually annotated dataset of only 100 images, which we release publicly as the first computer vision–ready benthic dataset from the Weddell Sea. The resulting model is capable of detecting a wide variety of benthic organisms, more than previous works in this area and to a higher level of granularity.

\section{Related Work}
\label{sec:related}

Early studies into machine-assisted in situ benthic biodiversity monitoring used local features and hand-engineered pipelines to identify specific organisms of interest \citep{dawkinsAutomaticScallopDetection2013, smithAutomatedCountingNorthern2007}. Such methods require extensive adaptation for new organisms or environments, limiting their use in broad biodiversity monitoring surveys enabled by advances in underwater imaging and affordable data storage.

Data-driven deep learning models capable of generalised, automated feature extraction, e.g., Convolutional Neural Networks (CNNs), have accelerated the pace of machine-assisted in situ benthic biodiversity monitoring research \citep{trotterSurveyingDeepReview2025}. Such models are often task-specific: image classifiers for taxonomic ID \citep{piechaudAutomatedIdentificationBenthic2019, zhouEchoAIDeeplearningBased2023}, object detectors for abundance estimation \citep{mariniLongTermAutomated2022, zhangMarineZoobenthosRecognition2024}, semantic segmenters for habitat mapping and behaviour analysis \citep{mizunoEfficientCoralSurvey2020, pavoniChallengesDeepLearningbased2021, harrisonMachineLearningApplications2021}, and instance segmenters for biomass estimation \citep{lutjensDeepLearningBased2021}.

Few studies focus specifically on the Antarctic benthos, likely due to its remoteness and high fieldwork costs. \citep{mariniLongTermAutomated2022} make use of a YOLOv5 \citep{jocherUltralyticsYolov5V702022} model pre-trained on the COCO dataset \citep{linMicrosoftCOCOCommon2014} to provide abundance estimates for two coarse-grained morphotypes, organism groupings classified based on shared morphological characteristics, in imagery from a stationary camera deployed in the Ross Sea. As imagery is downscaled, detection of small organisms may not be feasible using the presented methodology. 

\citep{lutjensDeepLearningBased2021} apply a patching strategy to towed-camera imagery from the Weddell Sea to evaluate the effectiveness of synthetic data augmentation in training a CenterMask \citep{leeCenterMaskRealTimeAnchorFree2020} model for instance segmentation of three coarse-grained morphotypes. However, their approach does not address potential detection failures at patch boundaries or support full-image ecological analysis. To address these limitations, we extend patching with overlap and postprocessing techniques that improve detection accuracy at patch edges. Additionally, we reproject detections back onto the original large-scale imagery, preserving spatial context for ecological analysis. Detailed methodology is provided in \cref{sec:methods}.

Further, while the aforementioned works demonstrate the feasibility of identifying Antarctic organisms using deep learning, they are limited to a small number of coarse-grained morphotypes. These studies also do not consider taxa with low abundance, potentially overlooking ecologically important but infrequently observed organisms. In contrast, we explore the development of object detection models that capture fine-grained taxonomic and morphological range, including rare organisms, examining the effect of abundance on model detection capability.
\section{The Weddell Sea Benthic Dataset}

\begin{figure}
	\begin{center}
		\includegraphics[width=\linewidth]{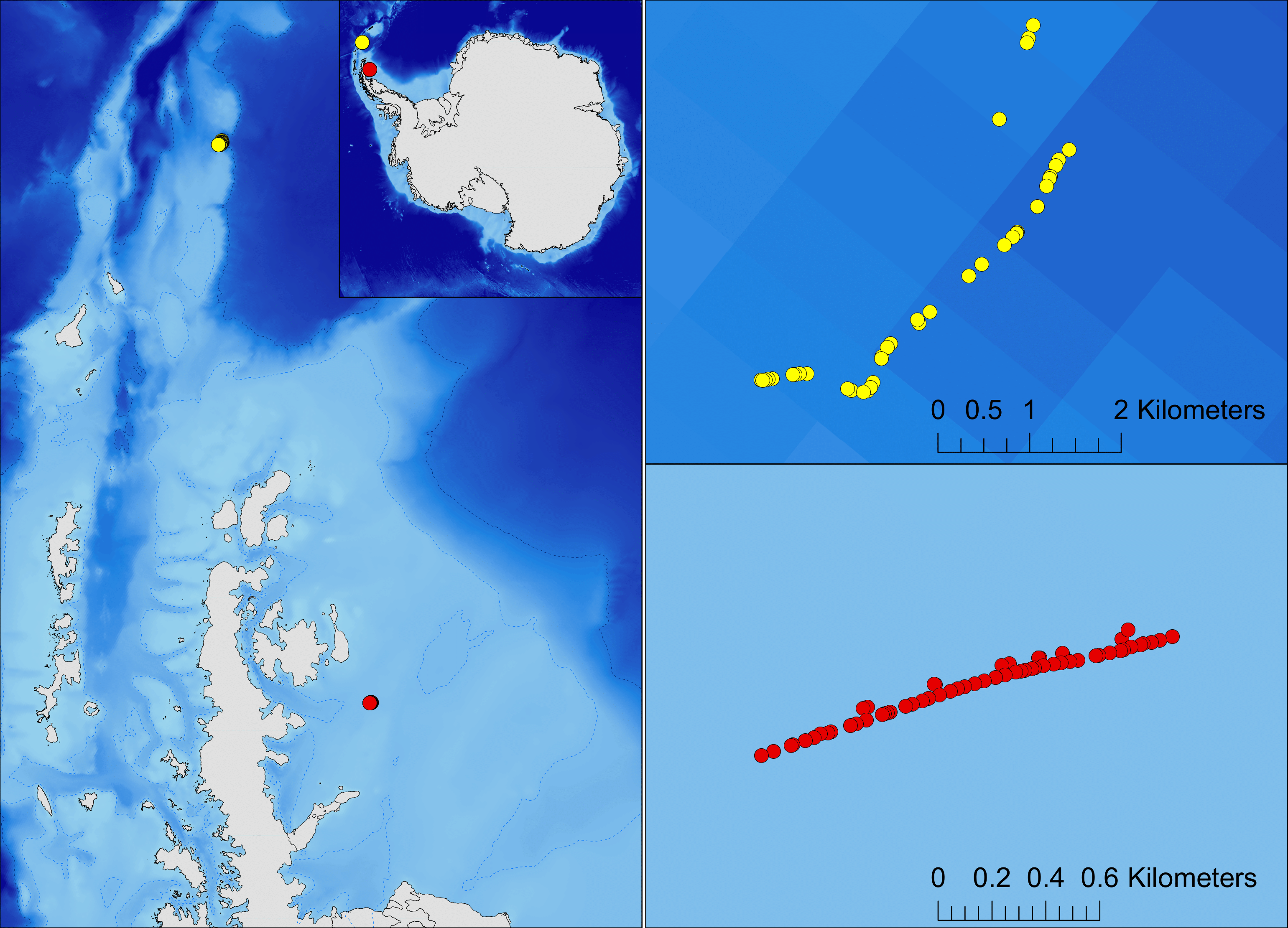}
	\end{center}
	\caption{Map of PS118 image acquisition sites included in the Weddell Sea Benthic Dataset. Yellow dots show data collected during cruise 69-1, while red points correspond to cruise 6-9.}
	\label{fig:datamap}
    \vspace{-2em}
\end{figure}

Data used in this study were collected during expedition PS118 (cruises 69-1 and 6-9; see \cref{fig:datamap}) of the RV \textit{Polarstern} \citep{purserSeabedVideoStill2021}. High-resolution benthic imagery (22 Megapixel, average filesize $=$ 6.94 Megabyte (MB)) was captured in a top-down view using the Ocean Floor Observation and Bathymetry System (OFOBS) \citep{purserOceanFloorObservation2019}, a towed camera system operating just above the seafloor. The imagery captures a diverse range of environmental conditions, including variable turbidity, illumination levels, and substrate types (hard and soft). Some images exhibit mild distortion due to the motion of the OFOBS during capture.

A subset of the collected imagery, selected for their ecological rather than model training merit, was manually annotated to facilitate benthic community composition analysis \citep{khanNetworkAnalysesPhotographic2024}; this forms the ground truth dataset used in this study, which we name The Weddell Sea Benthic Dataset (WSBD). This dataset comprises 100 annotated images captured at a range of depths (421--2202 m) and seafloor inclinations (0--80\degree). Where images were not comprehensively annotated, e.g., due to distortion, the unlabelled regions were cropped, resulting in images of varying sizes (average $=$ 3364×4545 px, 1.15 MB). Images are distinctly separated, with no overlap present between them. The original annotations were consolidated into 25 morphologically distinct classes, ranging from broad taxonomic groups to species level (see App. \ref{app:datasetClassStats}).

The dataset presents substantial visual complexity, with imagery characterised by high levels of background clutter, variable illumination, shadowing, and overlapping objects. These factors, plus the presence of fine-grained and morphologically similar taxa, make the WSBD a challenging and ecologically realistic benchmark for evaluating benthic object detection frameworks.

Imagery is biased towards soft-substrate environments at shallower depths (420--500 m), comprising 61.00\% of images but only 4.24\% of annotations. The dataset also shows a bias towards low-inclination areas, with 52.00\% of images taken on slopes $<=$10\degree. Certain taxa are restricted to specific substrate. 

Owing to the remoteness of the study site and limited anthropogenic disturbance under the Antarctic Treaty System, the WSDB contains high organism densities. The dataset contains 31,280 total bounding box annotations, with individual images containing between 5 and 1693 annotations (average $=$ 312.8). This includes numerous overlapping bounding boxes, a known challenge for object detection systems \citep{chattopadhyayCountingEverydayObjects2017, hoekendijkCountingUsingDeep2021}. Further, the dataset exhibits significant class imbalance, following a long-tailed distribution consistent with ecological patterns. The number of annotated instances per class ranges from 13,295 for stylasterids to 10 for the ascidian \textit{Cnemidocarpa verrucosa}. Addressing rare-class detection remains a critical issue in machine-assisted biodiversity monitoring \citep{vanhornDevilTailsFinegrained2017, vanhornINaturalistSpeciesClassification2018, miaoIterativeHumanAutomated2021}.

Small object detection remains a challenging and largely unsolved problem in computer vision \citep{LIU2021114602}. The WSBD dataset exemplifies these difficulties, exhibiting substantial inter-class size variation. Average bounding box areas range from 520 px$^{2}$ for cup corals to 68,092 px$^{2}$ for the ascidian \textit{Distaplia}. Further, intra-class size variability is introduced by fluctuations in the OFOBS' altitude during image capture, resulting in inconsistent scales which further complicate detection tasks. We release the WSBD under an OGL-UK-3.0 license: \url{https://doi.org/10.5285/1BA97E4B-EFB7-460B-9F2D-90437E33CE09}.
\section{Method}\label{sec:methods}

\begin{figure*}[t]
	\begin{center}
		\includegraphics[width=\linewidth]{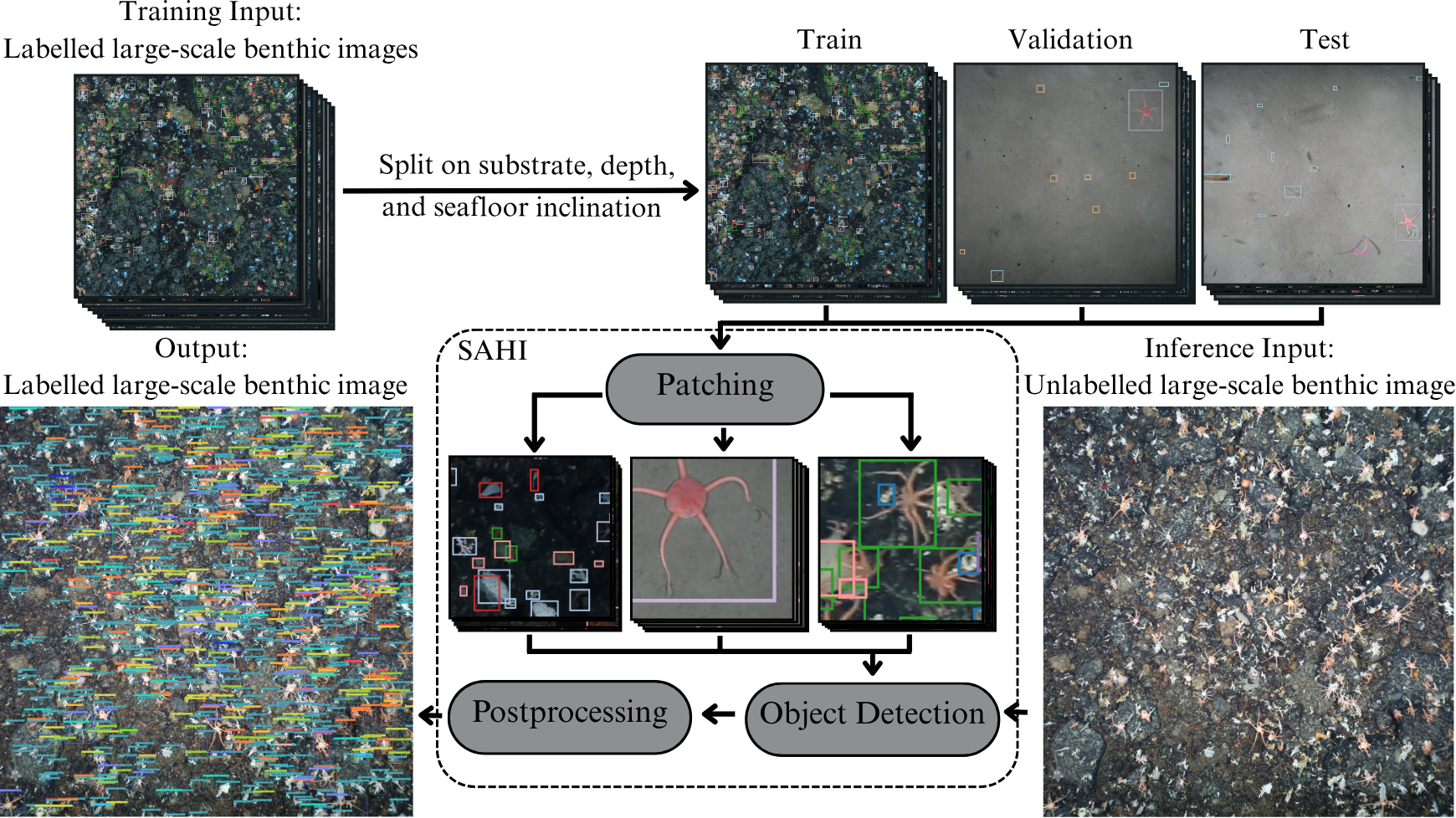}
	\end{center}
	\caption{A high level overview of the proposed Antarctic benthic organism detection and classification framework. For large-scale visualisation of the output, see App. \ref{app:exampleData}.}
	\label{fig:framework_overview}
\end{figure*}

Our proposed methodology (see \cref{fig:framework_overview}) enables us to exploit the high spatial fidelity of the WSBD whilst maintaining detection efficacy.

\subsection{Dataset Preparation}
To account for the imbalance in annotation volume between the two substrate types, the train, validation, and test sets were generated based on the proportion of total annotations rather than the number of images. These sets were then refined to ensure they remained representative of the geographic and environmental diversity present in the dataset, including variation in depth and seafloor inclination. This adjustment was made to enhance model generalisability and help prevent overfitting to specific environmental conditions, which is critical in biodiversity monitoring applications \citep{ottavianiAssessingImageConcept2022, wyattUsingEnsembleMethods2022}. The final annotation-level train–validation–test split was 68.71\%, 18.93\%, and 12.36\%, respectively.

\subsection{Image Patching}

The WSBD provides high-resolution benthic imagery which, while crucial for classifying small and morphologically similar organisms, introduces substantial computational overhead. Conventional object detection architectures are typically optimised for lower-resolution inputs \citep{NIPS2015_14bfa6bb, linFocalLossDense2017} and thus struggle to process full-resolution WSBD images without exceeding memory constraints. Downscaling such imagery to meet these limitations results in the loss of visual features which is particularly detrimental to the detection of small organisms (see \cref{sec:results_subsec:ablation_subsubsec:downscaling}).

To retain visual features we implement a patch-based detection strategy, subdividing the original large-scale image into sub-images of uniform size via a sliding window with fixed horizontal and vertical strides. Image patching is a well-established technique within machine-assisted in situ benthic biodiversity monitoring research, though it has primarily been employed for coarse-grained tasks \citep{gonzalez-riveroMonitoringCoralReefs2020, lutjensDeepLearningBased2021, jackettBenthicSubstrateClassification2023}. Patching also standardises input dimensions, mitigating image size variability from dataset generation and enabling more efficient training and inference.

To extend patch-based processing to object-level tasks, we adopt the Slicing Aided Hyper Inference (SAHI) methodology \citep{akyonSlicingAidedHyper2022}. Designed to improve the performance of object detection models on high-resolution data, SAHI works by dividing large images into overlapping patches, applying patch-based detection, and subsequently merging results via postprocessing. This approach retains resolution and aids small object detection, a common problem in ecologically complex imagery. SAHI has demonstrated strong performance in object-level tasks involving high-resolution data across other domains \citep{chaurasiaRealtimeDetectionBirds2023, muzammulEnhancingUAVAerial2024, giaEnhancingRoadObject2024}. We evaluate the optimal SAHI patching configuration, including patch size, overlap stride, and minimum bounding box visibility (the proportion of a ground truth bounding box required within a patch to be considered a valid object instance).

\subsection{Object Detection}
An object detection model is trained using patches as input, allowing for the retention of fine-grained features necessary for accurately detecting small and densely clustered organisms. The model generates bounding boxes around proposed regions of interest per patch, accompanied by a predicted class label, corresponding to one of 25 defined organism morphotypes, and a confidence score. We evaluate a range of object detection architectures, including single-stage and two-stage detectors as well as CNN and transformer-based models, alongside various data augmentation strategies and model fine-tuning.

\subsection{Postprocessing}
Following inference, patch-level detections are mapped back to their original coordinates on the large-scale input image. To resolve redundant detections resulting from overlapping patches, we apply a Non-Maximum Merging (NMM) procedure, consolidating multiple detections of the same object into a single bounding box.

\section{Experiments}\label{sec:experiments}

We evaluate various methodological configurations to determine the optimal framework setup for the WSBD. Specifically, we examine the impact of different SAHI parameters, augmentation strategies, architectures, and the use of pre-trained weights for model fine-tuning. Throughout, we examine the effect of organism abundance on model performance. The final optimal setup uncovered represents a baseline benchmark for the WSBD.

\subsection{Experimental Setup}

Experiments were run on a single High Performance Computing node using one NVIDIA A2 GPU. Object detection models were implemented in Python using MMDetection \citep{chenMMDetectionOpenMMLab2019}, with data augmentation via Albumentations \cite{buslaevAlbumentationsFastFlexible2020}. Training was performed for up to 200 epochs, with early stopping if no improvement was seen after 10 epochs. Our code is available at: \url{https://github.com/Trotts/antarctic-benthic-organism-detection/}.

Unless stated otherwise, all experiments used a Faster R-CNN architecture \citep{NIPS2015_14bfa6bb}. Evaluation used Mean Average Precision (mAP) across multiple Intersection over Union (IoU) thresholds and object sizes (Small, Medium, and Large), following the COCO format \citep{linMicrosoftCOCOCommon2014}. Models were evaluated on both the full 25-class set and a 10-class subset comprising the most abundant taxa. Metrics were computed after patch-level detections were reprojected back to their original large-scale image coordinates and postprocessed using NMM.

\subsection{SAHI Patching Parameters}

To implement SAHI effectively, several parameters must be defined to control how images and annotations are divided into patches. To determine the optimal configuration for the WSBD, we conducted a series of experiments training a model for each combination of three key parameters: patch size (250×250, 500×500, 750×750, and 1000×1000 px), overlap stride (0.0, 0.25, and 0.50), and minimum bounding box visibility (0.10, 0.25, and 0.50). The NMM IoU threshold was fixed at 0.5 across all configurations.

Evaluation revealed that a patch size of 500×500 px with a 0.50 stride and a minimum bounding box visibility of 0.25 achieved the highest overall performance (see \cref{tab:baseline_results}). This configuration maintained robust performance across all object sizes relative to other tested permutations. While larger patch sizes yielded comparable results for detecting larger organisms, they were less effective for smaller taxa, suggesting a trade-off between patch size and sensitivity to fine-grained features. Additionally, larger patches and higher stride increased computational demands due to increased dataset size and model input parameters. A minimum bounding box visibility of 0.25 yielded the highest mAP. Lower thresholds introduced training noise by retaining extremely cropped objects, while higher thresholds excluded valid examples, disproportionately affecting small organisms that frequently occur near patch boundaries.

This setup generated 25,184 patches from the 100 WSBD images, with 17,819 used for training, 4310 for validation, and 3055 for testing. 

\begin{table}[]
\caption{Test set Mean Average Precision (mAP) across key Intersection over Union (IoU) thresholds and object sizes for each dataset configuration. Bold indicates top performance per metric.}
\label{tab:baseline_results}
\resizebox{\columnwidth}{!}{%
\begin{tabular}{l|cccccccccc|}
\cline{2-11}
 &
  \multicolumn{10}{c|}{\textbf{Mean Average Precision (mAP)}} \\ \cline{2-11} 
 &
  \multicolumn{2}{c|}{\multirow{2}{*}{\textbf{@0.5:0.95}}} &
  \multicolumn{8}{c|}{\textbf{@0.5}} \\ \cline{4-11} 
 &
  \multicolumn{2}{c|}{} &
  \multicolumn{2}{c|}{\textbf{All}} &
  \multicolumn{2}{c|}{\textbf{Small}} &
  \multicolumn{2}{c|}{\textbf{Medium}} &
  \multicolumn{2}{c|}{\textbf{Large}} \\ \hline
\multicolumn{1}{|l|}{\diagbox{\textbf{Parameters}}{\textbf{Num. Classes}}} &
  \textbf{10} &
  \multicolumn{1}{c|}{\textbf{25}} &
  \textbf{10} &
  \multicolumn{1}{c|}{\textbf{25}} &
  \textbf{10} &
  \multicolumn{1}{c|}{\textbf{25}} &
  \textbf{10} &
  \multicolumn{1}{c|}{\textbf{25}} &
  \textbf{10} &
  \textbf{25} \\ \hline
\multicolumn{1}{|l|}{SAHI Patching} &
  \textbf{0.22} &
  \multicolumn{1}{c|}{0.18} &
  \textbf{0.45} &
  \multicolumn{1}{c|}{0.34} &
  0.20 &
  \multicolumn{1}{c|}{\textbf{0.24}} &
  0.48 &
  \multicolumn{1}{c|}{\textbf{0.35}} &
  \textbf{0.54} &
  0.42 \\
\multicolumn{1}{|l|}{+ SAHI Postprocessing} &
  0.21 &
  \multicolumn{1}{c|}{\textbf{0.19}} &
  \textbf{0.45} &
  \multicolumn{1}{c|}{\textbf{0.37}} &
  0.20 &
  \multicolumn{1}{c|}{0.23} &
  \textbf{0.50} &
  \multicolumn{1}{c|}{0.33} &
  0.52 &
  \textbf{0.44} \\
\multicolumn{1}{|l|}{+ Spatial Augmentation} &
  0.21 &
  \multicolumn{1}{c|}{0.18} &
  \textbf{0.45} &
  \multicolumn{1}{c|}{0.33} &
  \textbf{0.22} &
  \multicolumn{1}{c|}{0.19} &
  \textbf{0.50} &
  \multicolumn{1}{c|}{0.32} &
  0.49 &
  \textbf{0.44} \\ \hline
\end{tabular}%
}
\vspace{-1.5em}
\end{table}

\subsection{SAHI Postprocessing Parameters}\label{sec:results_subsec:postprocessingParameters}

During SAHI postprocessing, predicted patch-level bounding boxes are reprojected to their original coordinates within the large-scale input image. Bounding boxes of the same class that overlap by at least a specified IoU threshold are merged using NMM to reduce duplicate detections. To identify the optimal NMM IoU threshold, we applied SAHI postprocessing to the optimal patching model across a range of IoU values from 0.05 to 0.50 in 0.05 increments.

An IoU threshold of 0.20 was found to be sufficient, indicating a relatively low overlap threshold is optimal for merging duplicate detections after reprojection, particularly given the prevalence of small, densely clustered objects found within hard substrate imagery. Higher thresholds often failed to merge duplicate detections, resulting in inflated false positives, while lower thresholds erroneously merge distinct but nearby objects. 

\subsection{Data Augmentation Strategy}\label{sec:results_subsec:augmentation}

Given the limited volume of training data in the WSBD, we evaluated the effect of data augmentation, generating new samples by perturbing existing data, on model performance. Data scarcity is a persistent challenge for the development of machine-assisted in situ benthic biodiversity monitoring tools, though prior studies have shown that augmentation can enhance model performance \citep{durdenAutomatedClassificationFauna2021, pavoniChallengesDeepLearningbased2021, lutjensDeepLearningBased2021, doigDetectingEndangeredMarine2024}.

We evaluated three augmentation strategies: pixel-level, spatial-level, and a combined approach using both (see App. \ref{app:dataAug}). Each was assessed against a non-augmented baseline defined in \cref{sec:results_subsec:postprocessingParameters}. Spatial transformations yielded the best overall results. This is likely due to the WSBD's existing artefacts, such as motion blur and shadow, reducing the effectiveness of additional pixel-level perturbations. In contrast, spatial transformations introduced beneficial variability without further obscuring key visual features. This approach achieved the highest mAP@0.5 for small objects and tied for the best performance on medium objects in the 10-class setup. These improvements are particularly valuable in the WSBD, where small, densely packed organisms are frequent and difficult to detect. Accordingly, spatial augmentation was adopted for all subsequent experiments.

\subsection{Architecture Search}

To explore a broader range of detection capabilities beyond the Faster R-CNN baseline, we evaluated a mix of single-stage and two-stage detectors, as well as CNN and transformer-based architectures, against WSBD performance.

The DINO \citep{zhangDINODETRImproved2022} architecture achieved the highest mAP@0.5:0.95 for both the 10-class and 25-class configurations, sharing the top score with Cascade R-CNN \citep{cai2018cascade} in the latter (See \cref{tab:architecture-search} Top). When evaluating at mAP@0.5, DINO also outperformed other models in the 10-class setup, while Deformable-DETR \citep{zhuDeformableDETRDeformable2021} achieved the highest performance in the 25-class setting. While no architecture surpassed the Faster R-CNN baseline in detecting small and medium objects under the 10-class configuration, Co-DETR \citep{mengConditionalDETRFast2023} exhibited the best performance on small objects in the 25-class evaluation.

For medium-sized objects under the same configuration, the top-performing model was shared between Faster R-CNN and DINO. In the case of large object detection, RetinaNet \citep{linFocalLossDense2017} delivered the best results in the 10-class evaluation. However, its performance dropped substantially in the 25-class setting, where Cascade R-CNN significantly outperformed all other architectures.

\begin{table}[]
\caption{Test set Mean Average Precision (mAP) across key Inetrsection over Union (IoU) thresholds and object sizes for various model architectures (ordered by release), trained with optimal dataset settings, with and without COCO fine-tuning. Bold indicates top performance per metric; italics denote average performance.}
\label{tab:architecture-search}
\resizebox{\columnwidth}{!}{%
\begin{tabular}{cl|cccccccccc|}
\cline{3-12}
\multicolumn{1}{l}{} &
   &
  \multicolumn{10}{c|}{\textbf{Mean Average Precision (mAP)}} \\ \cline{3-12} 
\multicolumn{1}{l}{} &
   &
  \multicolumn{2}{c|}{\multirow{2}{*}{\textbf{@0.5:0.95}}} &
  \multicolumn{8}{c|}{\textbf{@0.5}} \\ \cline{5-12} 
\multicolumn{1}{l}{} &
   &
  \multicolumn{2}{c|}{} &
  \multicolumn{2}{c|}{\textbf{All}} &
  \multicolumn{2}{c|}{\textbf{Small}} &
  \multicolumn{2}{c|}{\textbf{Medium}} &
  \multicolumn{2}{c|}{\textbf{Large}} \\ \hline
\multicolumn{1}{|l|}{\textbf{Fine-tuned}} &
  \diagbox{\textbf{Architecture}}{\textbf{Num. Classes}} &
  \textbf{10} &
  \multicolumn{1}{c|}{\textbf{25}} &
  \textbf{10} &
  \multicolumn{1}{c|}{\textbf{25}} &
  \textbf{10} &
  \multicolumn{1}{c|}{\textbf{25}} &
  \textbf{10} &
  \multicolumn{1}{c|}{\textbf{25}} &
  \textbf{10} &
  \textbf{25} \\ \hline
\multicolumn{1}{|c|}{\multirow{6}{*}{\xmark}} &
  Faster R-CNN \citep{NIPS2015_14bfa6bb} &
  0.21 &
  \multicolumn{1}{c|}{0.18} &
  0.45 &
  \multicolumn{1}{c|}{0.33} &
  \textbf{0.22} &
  \multicolumn{1}{c|}{0.19} &
  0.50 &
  \multicolumn{1}{c|}{0.32} &
  0.49 &
  0.44 \\
\multicolumn{1}{|c|}{} &
  Cascade R-CNN \citep{cai2018cascade} &
  0.20 &
  \multicolumn{1}{c|}{\textbf{0.19}} &
  0.42 &
  \multicolumn{1}{c|}{0.35} &
  0.13 &
  \multicolumn{1}{c|}{0.10} &
  0.45 &
  \multicolumn{1}{c|}{0.30} &
  0.59 &
  \textbf{0.55} \\
\multicolumn{1}{|c|}{} &
  RetinaNet \citep{linFocalLossDense2017} &
  0.18 &
  \multicolumn{1}{c|}{0.12} &
  0.38 &
  \multicolumn{1}{c|}{0.24} &
  0.09 &
  \multicolumn{1}{c|}{0.11} &
  0.42 &
  \multicolumn{1}{c|}{0.24} &
  0.61 &
  0.35 \\
\multicolumn{1}{|c|}{} &
  Deformable-DETR \citep{zhuDeformableDETRDeformable2021} &
  0.19 &
  \multicolumn{1}{c|}{0.18} &
  0.45 &
  \multicolumn{1}{c|}{0.36} &
  0.21 &
  \multicolumn{1}{c|}{0.15} &
  0.47 &
  \multicolumn{1}{c|}{0.31} &
  0.52 &
  0.46 \\
\multicolumn{1}{|c|}{} &
  DINO \citep{zhangDINODETRImproved2022} &
  0.22 &
  \multicolumn{1}{c|}{\textbf{0.19}} &
  0.46 &
  \multicolumn{1}{c|}{0.35} &
  0.16 &
  \multicolumn{1}{c|}{0.14} &
  0.49 &
  \multicolumn{1}{c|}{0.32} &
  0.53 &
  0.44 \\
\multicolumn{1}{|c|}{} &
  CoDETR \citep{mengConditionalDETRFast2023} &
  0.19 &
  \multicolumn{1}{c|}{0.16} &
  0.45 &
  \multicolumn{1}{c|}{0.34} &
  0.18 &
  \multicolumn{1}{c|}{0.22} &
  0.48 &
  \multicolumn{1}{c|}{0.31} &
  0.52 &
  0.43 \\ \cline{2-12} 
\multicolumn{1}{|l|}{} &
  \textit{Average} &
  \multicolumn{1}{l}{\textit{0.20}} &
  \multicolumn{1}{l|}{\textit{0.17}} &
  \multicolumn{1}{l}{\textit{0.44}} &
  \multicolumn{1}{l|}{\textit{0.33}} &
  \multicolumn{1}{l}{\textit{0.17}} &
  \multicolumn{1}{l|}{\textit{0.15}} &
  \multicolumn{1}{l}{\textit{0.47}} &
  \multicolumn{1}{l|}{\textit{0.30}} &
  \multicolumn{1}{l}{\textit{0.54}} &
  \multicolumn{1}{l|}{\textit{0.45}} \\ \hline
\multicolumn{1}{|c|}{\multirow{7}{*}{\cmark}} &
  Faster R-CNN \citep{NIPS2015_14bfa6bb} &
  0.21 &
  \multicolumn{1}{c|}{0.18} &
  0.48 &
  \multicolumn{1}{c|}{0.36} &
  \textbf{0.22} &
  \multicolumn{1}{c|}{0.21} &
  0.49 &
  \multicolumn{1}{c|}{0.31} &
  0.56 &
  0.44 \\
\multicolumn{1}{|c|}{} &
  Cascade R-CNN \citep{cai2018cascade} &
  0.22 &
  \multicolumn{1}{c|}{0.17} &
  0.46 &
  \multicolumn{1}{c|}{0.34} &
  0.18 &
  \multicolumn{1}{c|}{0.18} &
  0.48 &
  \multicolumn{1}{c|}{0.29} &
  0.53 &
  0.45 \\
\multicolumn{1}{|c|}{} &
  RetinaNet \citep{linFocalLossDense2017} &
  0.20 &
  \multicolumn{1}{c|}{\textbf{0.19}} &
  0.41 &
  \multicolumn{1}{c|}{0.34} &
  0.11 &
  \multicolumn{1}{c|}{0.12} &
  0.46 &
  \multicolumn{1}{c|}{0.31} &
  0.50 &
  0.45 \\
\multicolumn{1}{|c|}{} &
  Deformable-DETR \citep{zhuDeformableDETRDeformable2021} &
  0.22 &
  \multicolumn{1}{c|}{\textbf{0.19}} &
  \textbf{0.49} &
  \multicolumn{1}{c|}{\textbf{0.39}} &
  0.21 &
  \multicolumn{1}{c|}{\textbf{0.27}} &
  \textbf{0.51} &
  \multicolumn{1}{c|}{\textbf{0.33}} &
  0.55 &
  0.47 \\
\multicolumn{1}{|c|}{} &
  DINO \citep{zhangDINODETRImproved2022} &
  \textbf{0.24} &
  \multicolumn{1}{c|}{\textbf{0.19}} &
  0.47 &
  \multicolumn{1}{c|}{0.33} &
  0.19 &
  \multicolumn{1}{c|}{0.13} &
  0.49 &
  \multicolumn{1}{c|}{0.29} &
  0.57 &
  0.49 \\
\multicolumn{1}{|c|}{} &
  CoDETR \citep{mengConditionalDETRFast2023} &
  0.22 &
  \multicolumn{1}{c|}{\textbf{0.19}} &
  0.46 &
  \multicolumn{1}{c|}{0.33} &
  0.17 &
  \multicolumn{1}{c|}{0.12} &
  0.49 &
  \multicolumn{1}{c|}{0.30} &
  \textbf{0.60} &
  0.44 \\ \cline{2-12} 
\multicolumn{1}{|c|}{} &
  \textit{Average} &
  \multicolumn{1}{l}{\textit{0.22}} &
  \multicolumn{1}{l|}{\textit{0.19}} &
  \multicolumn{1}{l}{\textit{0.46}} &
  \multicolumn{1}{l|}{\textit{0.35}} &
  \multicolumn{1}{l}{\textit{0.18}} &
  \multicolumn{1}{l|}{\textit{0.17}} &
  \multicolumn{1}{l}{\textit{0.49}} &
  \multicolumn{1}{l|}{\textit{0.31}} &
  \multicolumn{1}{l}{\textit{0.55}} &
  \multicolumn{1}{l|}{\textit{0.46}} \\ \hline
\end{tabular}%
}
\vspace{-1.5em}
\end{table}

\subsection{Effect of Fine-tuning}

Alongside data augmentation (see \cref{sec:results_subsec:augmentation}), fine-tuning is an effective strategy for improving model generalisability in object detection tasks where training data is limited \citep{baltrusaitisMultimodalMachineLearning2019}. Rather than initialising model weights randomly, requiring the network to learn fundamental visual representations from scratch, fine-tuned models leverage weights obtained from previous training on large-scale datasets. This facilitates transfer learning, wherein knowledge acquired in a source domain is applied to enhance performance in a target domain \citep{baltrusaitisMultimodalMachineLearning2019}.

Given the challenge of limited labelled data in the development of machine-assisted in situ benthic biodiversity monitoring tools, fine-tuning has become a standard approach within the field \citep{trotterSurveyingDeepReview2025}. Here, we evaluate the impact of fine-tuning on WSBD performance by comparing models initialised with random weights to those initialised on weights derived after training on the COCO dataset \citep{linMicrosoftCOCOCommon2014}.

\begin{figure}
    \centering
    \includegraphics[width=\linewidth]{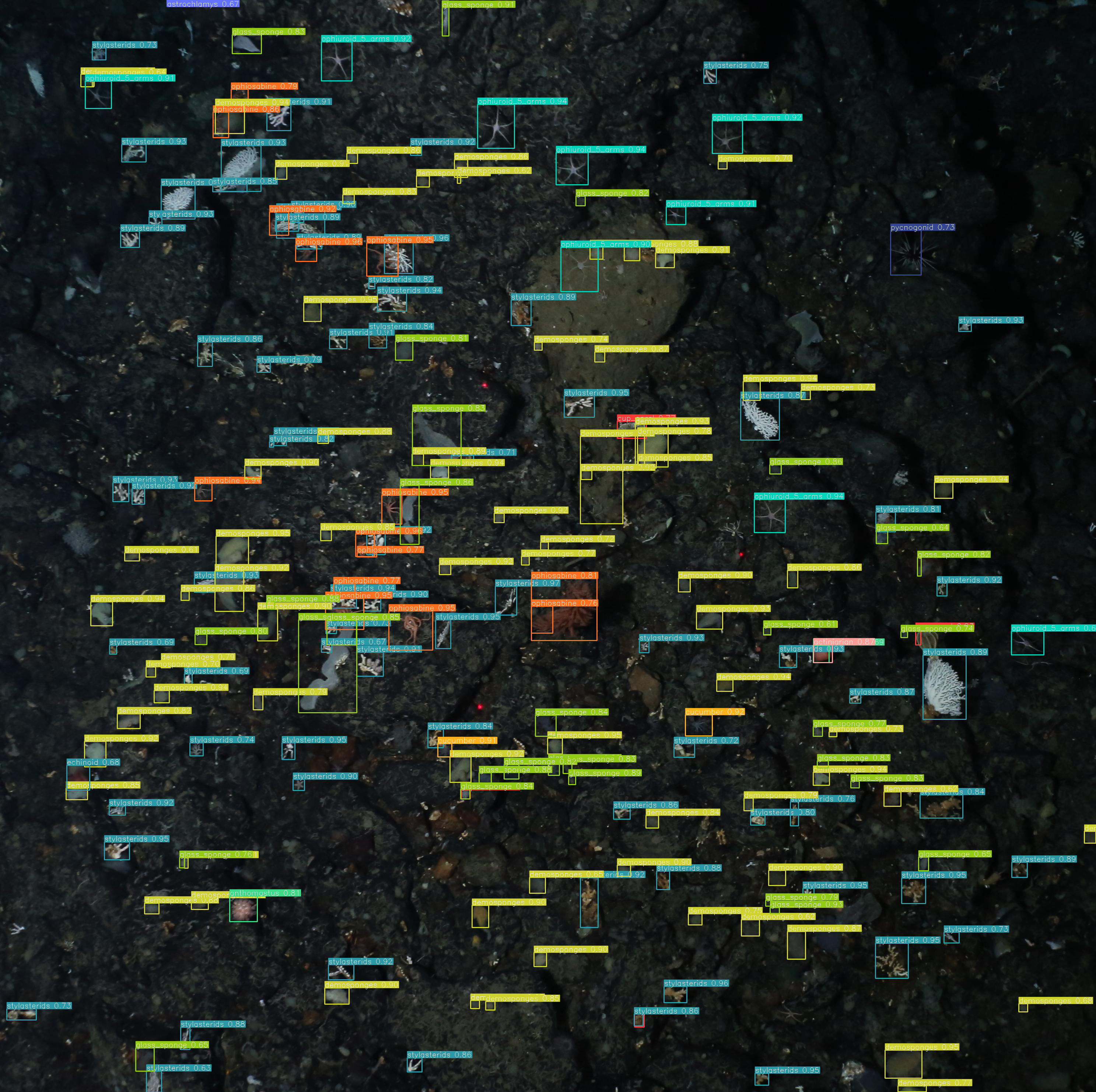}
    \caption{Example WSBD test set image output. Predicted organism bounding boxes, class labels, and confidence scores shown after reprojection and postprocessing. Confidence threshold $= 0.60$. For large-scale visualisations, see App. \ref{app:exampleData}.}
    \label{fig:exampleOut}
    \vspace{-1.5em}
\end{figure}

Overall, COCO fine-tuning resulted in a slight improvement in average model performance (see \cref{tab:architecture-search} Bottom). With the exception of large object detection in the 25-class evaluation, the highest metrics across evaluation categories were achieved by fine-tuned models. Notably, in contrast to the non-fine-tuned models where optimal performance varied depending on the specific evaluation scenario (e.g., object size or number of classes), the use of fine-tuning consistently elevated Deformable-DETR to either the top-performing model or among the top three performers across nearly all categories. This suggests that while the absolute gains from fine-tuning may be limited, the approach contributes to increased robustness and consistency in model performance. Crucially, it enables the identification of a single architecture, Deformable-DETR, as the most effective model overall, providing a clear candidate for subsequent deployment. An example output from this model can be seen in \cref{fig:exampleOut}.

\subsection{Ablation Study}

To further evaluate model performance and verify that each component of the proposed framework contributes positively, we conducted a series of additional ablation studies.

\subsubsection{Image-level Downscaling}\label{sec:results_subsec:ablation_subsubsec:downscaling}

Although the use of patching is intended to aid the detection of small objects, the results presented in \cref{tab:architecture-search} indicate that all evaluated model architectures continue to exhibit limited performance when detecting those present in the WSBD. To verify patching contributed positively, an additional Deformable-DETR model was trained using non-patched imagery, spatially augmented and fine-tuned on the COCO dataset, for comparison. To ensure consistent image size we downscale the data to 1635×1635 px, the smallest WSBD image.

Substantial declines in detection performance were observed across all object categories (see \cref{tab:ablation}), indicating a critical loss of discriminative features resulting from image downscaling. These findings reinforce the necessity of employing a patching strategy to preserve resolution and maintain object-level detail.

\begin{table}[]
\caption{Test set Mean Average Precision (mAP) across key Intersection over Union (IoU) thresholds and object sizes for each ablation study. The baseline model corresponds to the optimal Deformable-DETR configuration. Bold indicates top performance per metric.}
\label{tab:ablation}
\resizebox{\columnwidth}{!}{%
\begin{tabular}{l|cccccccccc|}
\cline{2-11}
 &
  \multicolumn{10}{c|}{\textbf{Mean Average Precision (mAP)}} \\ \cline{2-11} 
 &
  \multicolumn{2}{c|}{\multirow{2}{*}{\textbf{@0.5:0.95}}} &
  \multicolumn{8}{c|}{\textbf{@0.5}} \\ \cline{4-11} 
 &
  \multicolumn{2}{c|}{} &
  \multicolumn{2}{c|}{\textbf{All}} &
  \multicolumn{2}{c|}{\textbf{Small}} &
  \multicolumn{2}{c|}{\textbf{Medium}} &
  \multicolumn{2}{c|}{\textbf{Large}} \\ \hline
\multicolumn{1}{|l|}{\diagbox{\textbf{Experiment}}{\textbf{Num. Classes}}} &
  \textbf{10} &
  \multicolumn{1}{c|}{\textbf{25}} &
  \textbf{10} &
  \multicolumn{1}{c|}{\textbf{25}} &
  \textbf{10} &
  \multicolumn{1}{c|}{\textbf{25}} &
  \textbf{10} &
  \multicolumn{1}{c|}{\textbf{25}} &
  \textbf{10} &
  \textbf{25} \\ \hline
\multicolumn{1}{|l|}{Baseline} &
  \textbf{0.22} &
  \multicolumn{1}{c|}{\textbf{0.19}} &
  \textbf{0.49} &
  \multicolumn{1}{c|}{\textbf{0.39}} &
  \textbf{0.21} &
  \multicolumn{1}{c|}{\textbf{0.27}} &
  \textbf{0.51} &
  \multicolumn{1}{c|}{\textbf{0.33}} &
  \textbf{0.56} &
  \textbf{0.47} \\
\multicolumn{1}{|l|}{Image-level Downscaling} &
  0.06 &
  \multicolumn{1}{c|}{0.07} &
  0.13 &
  \multicolumn{1}{c|}{0.12} &
  0.03 &
  \multicolumn{1}{c|}{0.02} &
  0.12 &
  \multicolumn{1}{c|}{0.06} &
  0.19 &
  0.19 \\
\multicolumn{1}{|l|}{Non-Maximum Suppression} &
  0.21 &
  \multicolumn{1}{c|}{\textbf{0.19}} &
  \textbf{0.48} &
  \multicolumn{1}{c|}{0.34} &
  0.20 &
  \multicolumn{1}{c|}{0.20} &
  \textbf{0.51} &
  \multicolumn{1}{c|}{0.33} &
  0.53 &
  0.39 \\
\multicolumn{1}{|l|}{No Postprocessing} &
  \multicolumn{1}{l}{0.13} &
  \multicolumn{1}{l|}{0.13} &
  \multicolumn{1}{l}{0.26} &
  \multicolumn{1}{l|}{0.22} &
  \multicolumn{1}{l}{0.13} &
  \multicolumn{1}{l|}{0.16} &
  \multicolumn{1}{l}{0.27} &
  \multicolumn{1}{l|}{0.19} &
  \multicolumn{1}{l}{0.32} &
  \multicolumn{1}{l|}{0.29} \\ \hline
\end{tabular}%
}
\vspace{-1em}
\end{table}

\subsubsection{SAHI Postprocessing Algorithm}

Following patch-level detection, bounding boxes are reprojected to their corresponding large-scale image coordinates. To address duplicate detections resulting from patch overlap, boxes with identical class labels, overlapping with an IoU $>=0.20$, are postprocessed using NMM.  However, SAHI also allows for the use of Non-Maximum Suppression (NMS), where only the overlapping box with the highest confidence level is retained. Additionally, detections may be reprojected without any postprocessing applied. To verify merging was the correct approach, we evaluated the use of NMS and no postprocessing after reprojection. 

Substituting NMM with NMS resulted in either a slight decrease or no measurable improvement in detection performance, particularly under the more challenging 25-class evaluation (see \cref{tab:ablation}). The decline was most evident for both small and large object categories, where the baseline model employing merging demonstrated superior performance. Further, the use of no postprocessing significantly reduces performance across all evaluated metrics. These findings suggest that merging methods more effectively preserve localisation quality in cases where object instances are fragmented across overlapping patches.

\section{Discussion} \label{sec:discussion}

Based on the results presented in \cref{sec:experiments}, we propose an optimal framework configuration for the fine-grained detection of benthic organisms in high-resolution towed camera imagery from the Weddell Sea, Antarctica. The recommended approach involves subdividing large-scale images into 500×500 px patches with a 0.50 horizontal and vertical overlap stride, alongside a minimum bounding box visibility threshold of 0.25. Dataset splitting is stratified by substrate type, depth, and seafloor inclination to ensure geographic and environmental diversity. The resulting patches are spatially augmented and used to train a Deformable-DETR object detection model, with initial weights derived from the COCO dataset. Following inference, detections are reprojected to their original locations on the full-resolution image. Overlapping same-class bounding boxes are then postprocessed using NMM with an IoU threshold of 0.20.

\subsection{Small Object Detection}

The resulting model is trained to detect 25 distinct morphotypes found in the Weddell Sea. However, notable performance limitations for small organisms are present, even when employing the SAHI methodology. While these limitations may partly stem from the restricted size of the training dataset, a common constraint in machine-assisted in situ benthic biodiversity monitoring, they are likely exacerbated by the logistical and environmental challenges of data collection in Antarctica.

The observed underperformance for small object detection, despite the use of high-resolution imagery, advanced patching strategies, data augmentation, and fine-tuning, suggests current object detection architectures are limited in their ability to extract meaningful features from small instances in visually complex benthic environments. This restricts accurate learning and detection of ecologically important taxa, and highlights the need for new architectural approaches tailored to small object representation. 

\subsection{Effect of Abundance and Morphology}\label{sec:discussion_Subsec:abundance}

Additionally, organism abundance was found to have a notable influence on model performance. This is evident when comparing the results of the 10-class and 25-class evaluations. The average number of annotations per class in the 10-class evaluation is 2068.5. In contrast, overall average abundance for the 25-class configuration is 859.4, dropping to just 53.4 for organisms present in the 25-class set only. Examining the optimal model's class confusion reveals that although the overall number of missed detections is high, especially for rare organisms, the rate of misclassification among detected abundant instances is low (see App. \ref{app:conf_mat}). This suggests that when the model places a bounding box, it is likely to contain a valid organism and to assign it the correct label. 

Where misclassification does occur it is typically between morphologically similar organisms, e.g., demosponges and glass sponges, which share structural features and can be difficult to distinguish visually, even for trained experts. In contrast, misclassifications between taxonomically related but visually distinct organisms, e.g., \textit{Ophiosabine} and other ophiuroids, are relatively rare. This indicates that the model relies primarily on visual cues rather than taxonomic proximity when assigning class labels. Interestingly, we observe strong detection performance for pycnogonids, despite this class being the fourth least abundant. However this may be due to a lack of morphological variation between the dataset splits for this class.

\subsection{SAHI Postprocessing Limitations}\label{sec:discussion_Subsec:postprocessingFails}

Unlike domains where SAHI is commonly applied, e.g., satellite imagery \citep{chaurasiaRealtimeDetectionBirds2023, muzammulEnhancingUAVAerial2024, giaEnhancingRoadObject2024}, data in the WSBD is captured from varying altitudes above the seafloor due to changes in OFOBS platform depth and seafloor topography. This introduces significant intra-class size variation. For large organisms, a single instance may span many patches. Experimental observations show SAHI occasionally struggles to accurately postprocess duplicate detections into a single coherent bounding box when an organism is divided across a large number of patches (see App. \ref{app:postprocessfail}). This negatively affects overall model performance and may bias abundance estimates if not addressed during manual post-hoc review.

\subsection{Expert Labelling Agreement}

It is important to note that evaluation metrics reported in this study, as in other automated biodiversity monitoring research, e.g., \citep{boulentScalingWhaleMonitoring2023}, reflect the degree of agreement between the model and the human annotator rather than an absolute measure of detection accuracy. Given the high densities of organisms, the prevalence of small-bodied taxa, and the well-documented issues of fatigue and subjectivity in manual annotation processes for benthic imagery \citep{culverhouseHumanMachineFactors2007, durdenComparisonImageAnnotation2016, piechaudAutomatedIdentificationBenthic2019}, it is likely some valid organisms were omitted from the ground truth, leading to correct model detections being penalised and artificially lowering performance metrics.

Due to the high level of taxonomic expertise required to accurately annotate Antarctic benthic fauna and the significant time investment needed (averaging approximately eight hours per image), it was not feasible to obtain multiple independent expert annotations to reduce potential labelling bias. With the time savings afforded by our framework, future labelling can incorporate consensus agreement.

\subsection{Potential Framework Application}

Despite these challenges, the resulting model remains highly valuable to benthic ecologists. The proposed framework offers the potential for substantial time and cost savings, particularly when applied to the processing of extensively backlogged survey data, totalling in the tens of thousands of images. As a result, the framework is well-suited for use in first-pass, human-in-the-loop analyses, allowing ecologists to focus on completing remaining annotations instead of reviewing full images manually.

A promising direction for future work involves integrating the proposed framework into an active learning pipeline, wherein unlabelled imagery is prioritised for annotation based on predefined selection criteria, automatically annotated using the framework, then refined by expert ecologists. Selection strategies may incorporate both ecological relevance and expected contribution to framework performance. As more archival data is processed through this iterative approach, the resulting enlarged dataset could serve as a valuable fine-tuning resource, especially for currently rare organisms where model performance may benefit from increased abundance, enabling iterative improvements in model performance as annotation efforts progress.
\section{Conclusion}
 
We address the challenge of detecting and classifying Antarctic benthic organisms in high-resolution, top-down imagery captured using a towed camera system in the Weddell Sea. Through the creation of the first publicly available computer vision–ready dataset of Antarctic seafloor ecology, we develop and assess a comprehensive object detection framework specifically designed for the complexities of benthic imagery. The proposed pipeline integrates the SAHI methodology, patching to retain spatial resolution and reduce computational expense, alongside spatial data augmentation and model fine-tuning to support generalisability under data-scarce conditions. Postprocessing using NMM enhances detection coherence after bounding box reprojection from patch-level back to the original large-scale image. Our framework demonstrates strong performance in detecting medium and large benthic morphotypes.

Persistent underperformance on small and rare taxa, even when enhancement strategies were applied, highlights fundamental limitations in current object detection architectures when applied to ecologically complex imagery. These findings underscore the need for targeted research into small object representation, as well as the potential value of active learning approaches to enable faster processing of backlogged, unprocessed field imagery, refining rare organism performance while uncovering new ecological insights. By providing our data and models open-source, we hope to encourage community efforts to improve the detection of such organisms in complex marine imagery. Nevertheless, our proposed framework offers a scalable, generalised solution for automated analysis of high-resolution benthic imagery, with significant potential to accelerate biodiversity monitoring and enable better protection of the unique benthic ecosystems found in Antarctica and beyond.

\section*{Acknowledgements}

We thank Miao Fan, Alfred Wegener Institutue (AWI), for providing the bathymetry maps used to subset the WSBD by seafloor inclination. We also thank Autun Purser, AWI, and all crew of the RV \textit{Polarstern} PS118 cruise for their data collection efforts. CT, HJG and RJW are funded by the UKRI Future Leaders Fellowship MR/W01002X/1 `The past, present and future of unique cold-water benthic (sea floor) ecosystems in the Southern Ocean' awarded to RJW. For the purpose of open access, the author(s) has applied a Creative Commons Attribution (CC BY) license to any Accepted Manuscript version arising.

{
    \small
    \bibliographystyle{ieeenat_fullname}
    \balance 
    \bibliography{main}
}

\clearpage 
\setcounter{page}{1}
\setcounter{section}{0}
\setcounter{figure}{0}
\setcounter{table}{0}

\let\oldthefigure\thefigure
\renewcommand{\thefigure}{S\oldthefigure}%
\let\oldthetable\thetable
\renewcommand{\thetable}{S\oldthetable}%

\renewcommand{\thesection}{\Alph{section}} 

\onecolumn
{
\newpage
   \centering
   \Large
   \textbf{\thetitle}\\
   \vspace{0.5em}Supplementary Material \\
   \vspace{1.0em}
}

\section{Dataset Composition}\label{app:datasetClassStats}

\begin{table}[H]
\caption{Counts and areas for the Weddell Sea Benthic Dataset classes, for both the original large-scale and patched images. Bolded counts denote the most abundant classes used for 10-class evaluation.}
\label{tab:datasetClassStats}
\resizebox{\textwidth}{!}{%
\begin{tabular}{l|lllllll|lllllll}
\cline{2-15}
 &
  \multicolumn{7}{c|}{\textbf{Whole Image Dataset}} &
  \multicolumn{7}{c|}{\textbf{Patched Dataset}} \\ \cline{2-15} 
 &
  \multicolumn{4}{c|}{\textbf{Count}} &
  \multicolumn{3}{c|}{\textbf{Area (px${^2}$)}} &
  \multicolumn{4}{c|}{\textbf{Count}} &
  \multicolumn{3}{c|}{\textbf{Area (px${^2}$)}} \\ \hline
\multicolumn{1}{|l|}{\textbf{Class Label}} &
  \textbf{All} &
  \textbf{Train} &
  \textbf{Validation} &
  \multicolumn{1}{l|}{\textbf{Test}} &
  \textbf{Min} &
  \textbf{Max} &
  \textbf{Avg} &
  \textbf{All} &
  \textbf{Train} &
  \textbf{Validation} &
  \multicolumn{1}{l|}{\textbf{Test}} &
  \textbf{Min} &
  \textbf{Max} &
  \multicolumn{1}{l|}{\textbf{Avg}} \\ \hline
\multicolumn{1}{|l|}{\texttt{actiniarian}} & 165 & 118 & 22 & \multicolumn{1}{l|}{25} & 111 & 51328 & 6821 & 777 & 553 & 105 & \multicolumn{1}{l|}{119} & 111 & 51328 & \multicolumn{1}{l|}{5512} \\
\multicolumn{1}{|l|}{\texttt{alcyonium}} & 280 & 266 & 6 & \multicolumn{1}{l|}{8} & 266 & 104355 & 6773 & 1269 & 1202 & 24 & \multicolumn{1}{l|}{43} & 60 & 98203 & \multicolumn{1}{l|}{5508} \\
\multicolumn{1}{|l|}{\texttt{anthomastus}} & 89 & 60 & 22 & \multicolumn{1}{l|}{7} & 198 & 30478 & 4666 & 371 & 260 & 80 & \multicolumn{1}{l|}{31} & 106 & 30478 & \multicolumn{1}{l|}{3810} \\
\multicolumn{1}{|l|}{\texttt{ascidian\_cnemidocarpa\_verrucosa}} & 10 & 2 & 4 & \multicolumn{1}{l|}{4} & 4689 & 91344 & 21113 & 55 & 15 & 20 & \multicolumn{1}{l|}{20} & 1517 & 91344 & \multicolumn{1}{l|}{15311} \\
\multicolumn{1}{|l|}{\texttt{ascidian\_distaplia}} & 32 & 24 & 1 & \multicolumn{1}{l|}{7} & 1303 & 1432444 & 68092 & 186 & 146 & 4 & \multicolumn{1}{l|}{36} & 539 & 250000 & \multicolumn{1}{l|}{34927} \\
\multicolumn{1}{|l|}{\texttt{ascidian\_pyura\_bouvetensis}} & 66 & 41 & 21 & \multicolumn{1}{l|}{4} & 981 & 188218 & 16652 & 351 & 203 & 126 & \multicolumn{1}{l|}{22} & 275 & 153083 & \multicolumn{1}{l|}{12022} \\
\multicolumn{1}{|l|}{\texttt{asteroidia}} & 156 & 111 & 23 & \multicolumn{1}{l|}{22} & 144 & 569242 & 9572 & 643 & 482 & 74 & \multicolumn{1}{l|}{87} & 26 & 250000 & \multicolumn{1}{l|}{8279} \\
\multicolumn{1}{|l|}{\texttt{astrochlamys}} & \textbf{720} & 528 & 127 & \multicolumn{1}{l|}{65} & 475 & 136010 & 13402 & \textbf{3366} & 2439 & 556 & \multicolumn{1}{l|}{371} & 110 & 126777 & \multicolumn{1}{l|}{10608} \\
\multicolumn{1}{|l|}{\texttt{benthic\_fish}} & 71 & 52 & 14 & \multicolumn{1}{l|}{5} & 309 & 179118 & 40603 & 474 & 349 & 92 & \multicolumn{1}{l|}{33} & 309 & 139500 & \multicolumn{1}{l|}{23828} \\
\multicolumn{1}{|l|}{\texttt{bryozoan}} & 15 & 8 & 5 & \multicolumn{1}{l|}{2} & 386 & 45004 & 14134 & 88 & 45 & 35 & \multicolumn{1}{l|}{8} & 386 & 45004 & \multicolumn{1}{l|}{9918} \\
\multicolumn{1}{|l|}{\texttt{crinoid}} & 26 & 19 & 3 & \multicolumn{1}{l|}{4} & 491 & 20989 & 6422 & 119 & 78 & 14 & \multicolumn{1}{l|}{27} & 232 & 20989 & \multicolumn{1}{l|}{5377} \\
\multicolumn{1}{|l|}{\texttt{crustaceans}} & \textbf{461} & 338 & 79 & \multicolumn{1}{l|}{44} & 554 & 399481 & 12799 & \textbf{2469} & 1817 & 424 & \multicolumn{1}{l|}{228} & 253 & 173560 & \multicolumn{1}{l|}{9494} \\
\multicolumn{1}{|l|}{\texttt{cucumber}} & \textbf{355} & 287 & 41 & \multicolumn{1}{l|}{27} & 62 & 9041 & 1522 & \textbf{1447} & 1166 & 164 & \multicolumn{1}{l|}{117} & 56 & 9041 & \multicolumn{1}{l|}{1415} \\
\multicolumn{1}{|l|}{\texttt{cup\_coral}} & \textbf{4757} & 2807 & 1611 & \multicolumn{1}{l|}{339} & 31 & 13756 & 520 & \textbf{18552} & 11262 & 6039 & \multicolumn{1}{l|}{1251} & 12 & 13756 & \multicolumn{1}{l|}{482} \\
\multicolumn{1}{|l|}{\texttt{demosponges}} & \textbf{2211} & 1517 & 340 & \multicolumn{1}{l|}{354} & 7 & 258424 & 2283 & \textbf{8960} & 6216 & 1312 & \multicolumn{1}{l|}{1432} & 7 & 194988 & \multicolumn{1}{l|}{2003} \\
\multicolumn{1}{|l|}{\texttt{echinoid}} & 11 & 6 & 2 & \multicolumn{1}{l|}{3} & 1975 & 12170 & 4098 & 50 & 24 & 12 & \multicolumn{1}{l|}{14} & 817 & 12170 & \multicolumn{1}{l|}{3607} \\
\multicolumn{1}{|l|}{\texttt{glass\_sponge}} & \textbf{2308} & 1612 & 477 & \multicolumn{1}{l|}{219} & 29 & 92647 & 1603 & \textbf{9144} & 6295 & 1930 & \multicolumn{1}{l|}{919} & 13 & 92647 & \multicolumn{1}{l|}{1508} \\
\multicolumn{1}{|l|}{\texttt{gorgonian}} & \textbf{1144} & 903 & 113 & \multicolumn{1}{l|}{128} & 62 & 303363 & 9045 & \textbf{5396} & 4248 & 546 & \multicolumn{1}{l|}{602} & 62 & 219445 & \multicolumn{1}{l|}{7274} \\
\multicolumn{1}{|l|}{\texttt{hydroid\_solitary}} & 25 & 17 & 5 & \multicolumn{1}{l|}{3} & 2519 & 97165 & 14968 & 133 & 94 & 25 & \multicolumn{1}{l|}{14} & 378 & 97165 & \multicolumn{1}{l|}{11492} \\
\multicolumn{1}{|l|}{\texttt{ophiosabine}} & \textbf{3075} & 1853 & 694 & \multicolumn{1}{l|}{528} & 219 & 50859 & 3125 & \textbf{12783} & 7630 & 2897 & \multicolumn{1}{l|}{2256} & 68 & 50859 & \multicolumn{1}{l|}{2768} \\
\multicolumn{1}{|l|}{\texttt{ophiuroid\_5\_arms}} & \textbf{1885} & 1293 & 280 & \multicolumn{1}{l|}{312} & 93 & 748890 & 17419 & \textbf{8819} & 5961 & 1349 & \multicolumn{1}{l|}{1509} & 72 & 250000 & \multicolumn{1}{l|}{13686} \\
\multicolumn{1}{|l|}{\texttt{pencil\_urchin}} & 78 & 51 & 20 & \multicolumn{1}{l|}{7} & 700 & 36549 & 6419 & 338 & 219 & 86 & \multicolumn{1}{l|}{33} & 130 & 36549 & \multicolumn{1}{l|}{5296} \\
\multicolumn{1}{|l|}{\texttt{pycnogonid}} & 11 & 7 & 2 & \multicolumn{1}{l|}{2} & 2381 & 25210 & 13854 & 59 & 40 & 9 & \multicolumn{1}{l|}{10} & 747 & 25210 & \multicolumn{1}{l|}{9264} \\
\multicolumn{1}{|l|}{\texttt{stylasterids}} & \textbf{13295} & 9547 & 2011 & \multicolumn{1}{l|}{1736} & 4 & 74999 & 1720 & \textbf{53523} & 38529 & 7543 & \multicolumn{1}{l|}{7451} & 4 & 69524 & \multicolumn{1}{l|}{1560} \\
\multicolumn{1}{|l|}{\texttt{worm\_tubes}} & 35 & 19 & 2 & \multicolumn{1}{l|}{14} & 168 & 25355 & 4591 & 157 & 82 & 8 & \multicolumn{1}{l|}{67} & 168 & 25355 & \multicolumn{1}{l|}{3958} \\ \hline
\multicolumn{1}{|l|}{\textit{Total}} &
  \textit{31280} &
  \textit{21486} &
  \textit{5925} &
  \multicolumn{1}{l|}{\textit{3869}} &
  \multicolumn{1}{c}{\textit{}} &
  \multicolumn{1}{c}{\textit{}} &
  \multicolumn{1}{c|}{\textit{}} &
  \textit{129529} &
  \textit{89355} &
  \textit{23474} &
  \multicolumn{1}{l|}{\textit{16700}} &
  \multicolumn{1}{c}{\textit{}} &
  \multicolumn{1}{c}{\textit{}} &
  \multicolumn{1}{c}{\textit{}} \\ \cline{1-5} \cline{9-12}
\end{tabular}%
}
\end{table}

\newpage

\section{Data Augmentation Strategy Details}\label{app:dataAug}
\cref{tab:dataAugs} presents the Albumentations-based \cite{buslaevAlbumentationsFastFlexible2020} data augmentation techniques used in this study, categorised by the augmentation strategy in which they were employed. Probabilities for all augmentations were set to 0.50. For \textit{Random Sized BBox Safe Crop}, the height and width parameters were set to the patch size. All other parameters were set to the Albumentations default.

Despite its name, \textit{Pixel Dropout} is classified as a spatial transformation. This augmentation operates non-uniformly across the image by randomly selecting specific spatial coordinates at which to drop pixels. Consequently, it alters the spatial structure of the image rather than applying a uniform change across all pixels.

\begin{table}[H]
\centering
\caption{A list of data augmentation techniques provided by the Albumentations library, along with the augmentation strategy in which each technique was applied.}
\label{tab:dataAugs}
\begin{tabular}{|l|ccc|}
\hline
\diagbox{\textbf{Augmentation}}{\textbf{Strategy}} & \textbf{Pixel} & \textbf{Spatial} & \textbf{Both} \\ \hline
Horizontal Flip                & \xmark & \cmark & \cmark \\
Motion Blur                    & \cmark & \xmark & \cmark \\
Pixel Dropout         & \xmark & \cmark & \cmark \\
Random Brightness and Contrast & \cmark & \xmark & \cmark \\
Random Shadow                  & \cmark & \xmark & \cmark \\
Random Sized BBox Safe Crop    & \xmark & \cmark & \cmark \\
Vertical Flip                  & \xmark & \cmark & \cmark \\ \hline
\end{tabular}
\end{table}

\newpage

\section{Weddell Sea Benthic Dataset High-Resolution Examples}\label{app:exampleData}

Example Weddell Sea Benthic Dataset test set image outputs. Predicted organism bounding boxes, class labels, and confidence scores shown after reprojection and postprocessing. Confidence threshold $=$ 0.60.

\begin{figure}[H]
	\begin{center}
    \caption{HOTKEY\_2019\_03\_31\_at\_13\_30\_13\_IMG\_0853. Original size: 3799×3798 px.}
	\includegraphics[width=\linewidth]{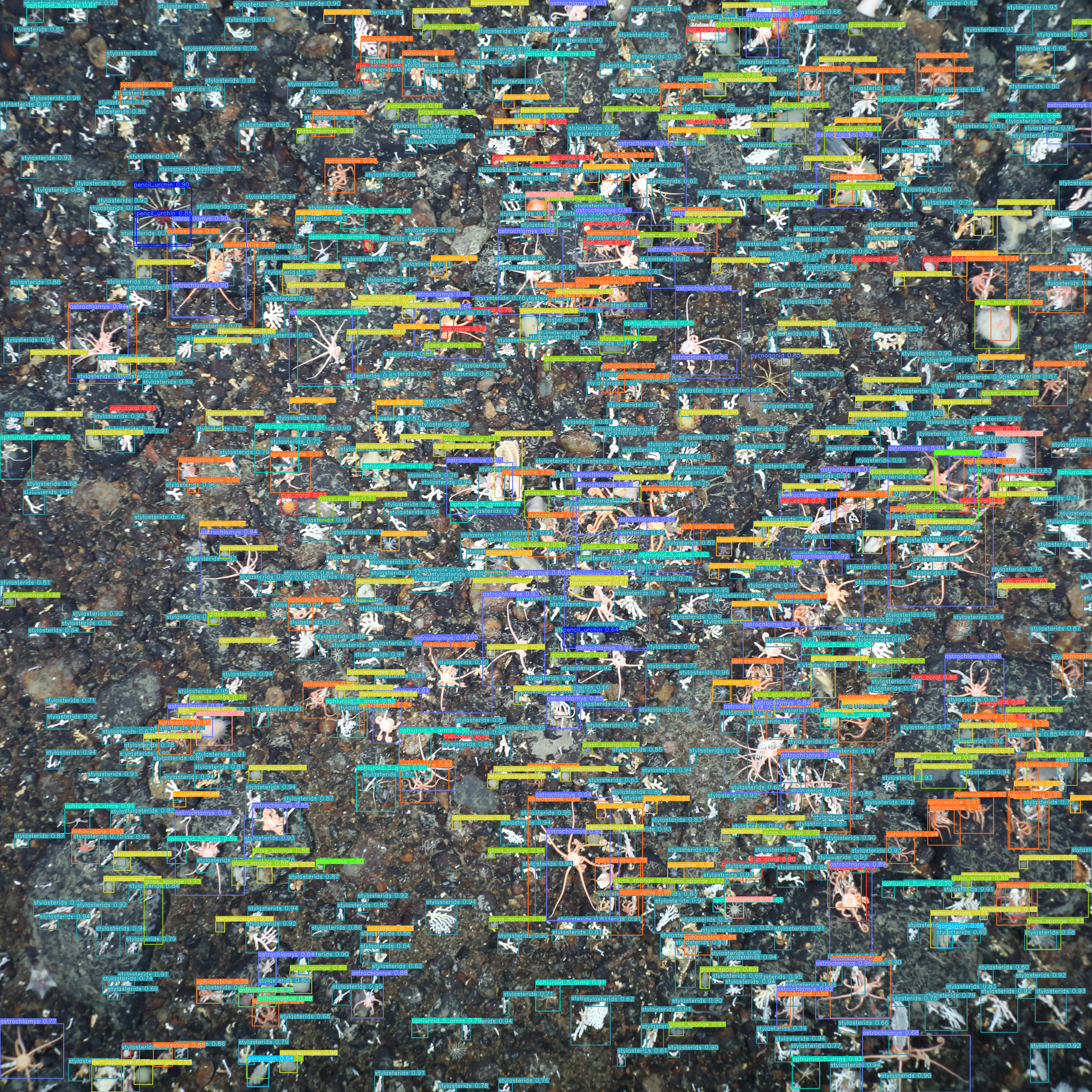}
	\end{center}
\end{figure}

\begin{figure}[H]
	\begin{center}
    \caption{HOTKEY\_2019\_03\_31\_at\_13\_21\_23\_IMG\_0816. Original size: 2975×2964 px.}
	\includegraphics[width=\linewidth]{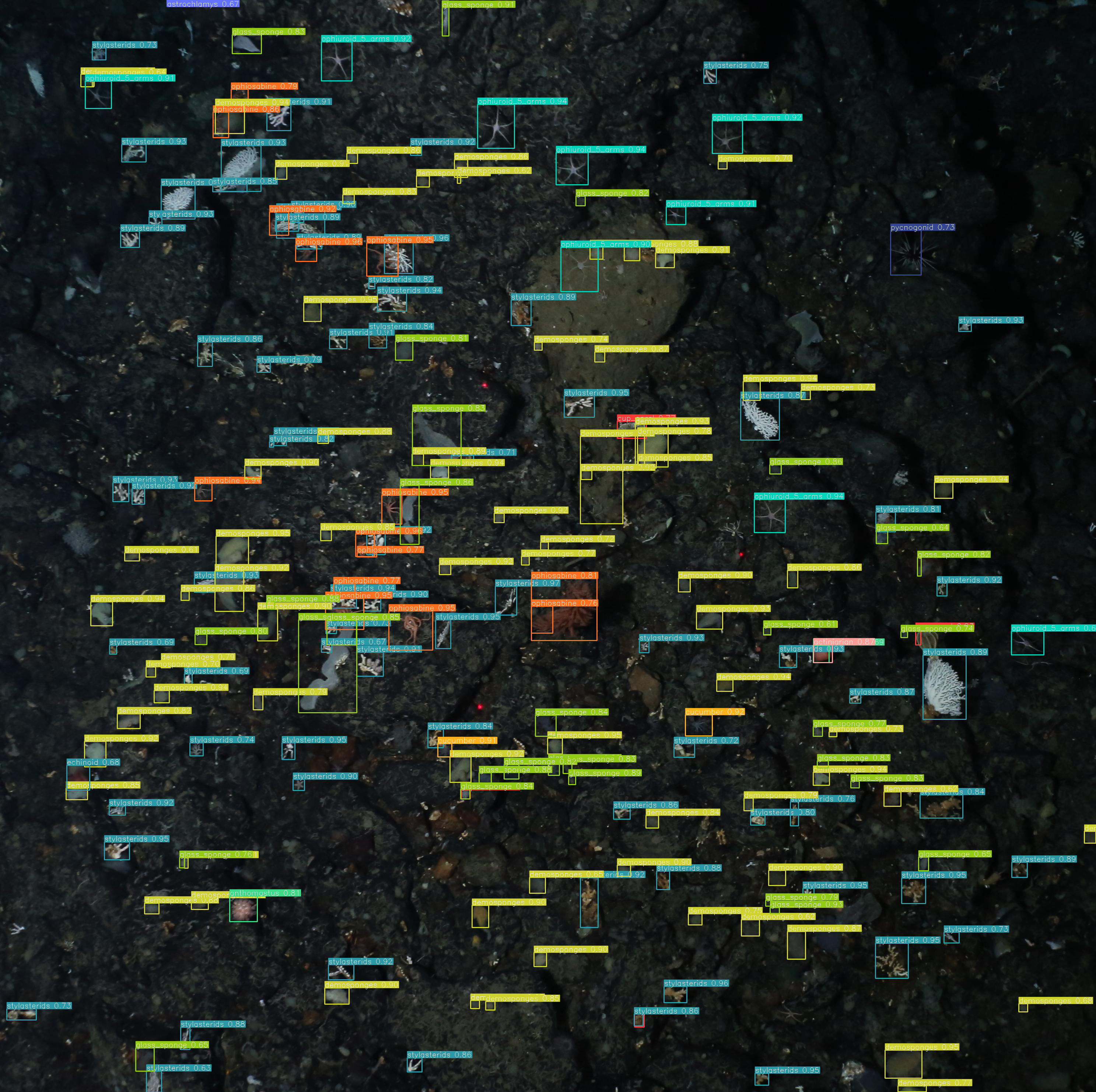}
	\end{center}
\end{figure}

\begin{landscape}
\begin{figure}[H]
	\begin{center}
    \caption{TIMER\_2019\_03\_06\_at\_05\_40\_47\_IMG\_0253. Original size: 5760×3840 px.}
	\includegraphics[width=\linewidth]{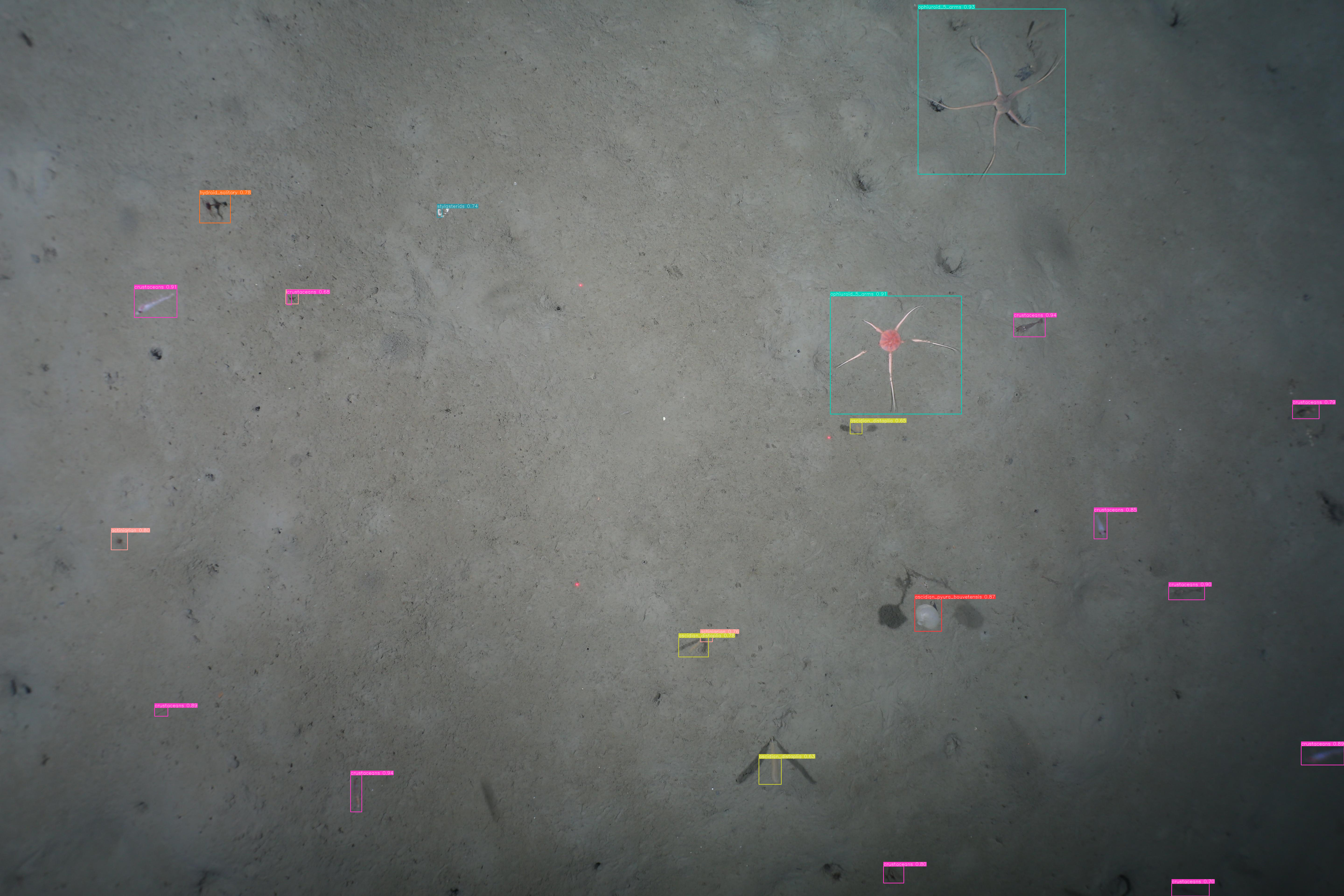}
	\end{center}
\end{figure}
\end{landscape}
\newpage

\section{Effect of Abundance on Model Performance}\label{app:conf_mat}

Organism abundance was shown to have a large effect on framework performance. See \ref{sec:discussion_Subsec:abundance} for further discussion.

\begin{figure}[H]
    \centering
    \caption{Confusion matrix for the optimal framework configuration, ordered by abundance. Red lines indicate the top-10 most abundant classes. Confidence threshold $=$ 0.60.}
    \includegraphics[width=\linewidth]{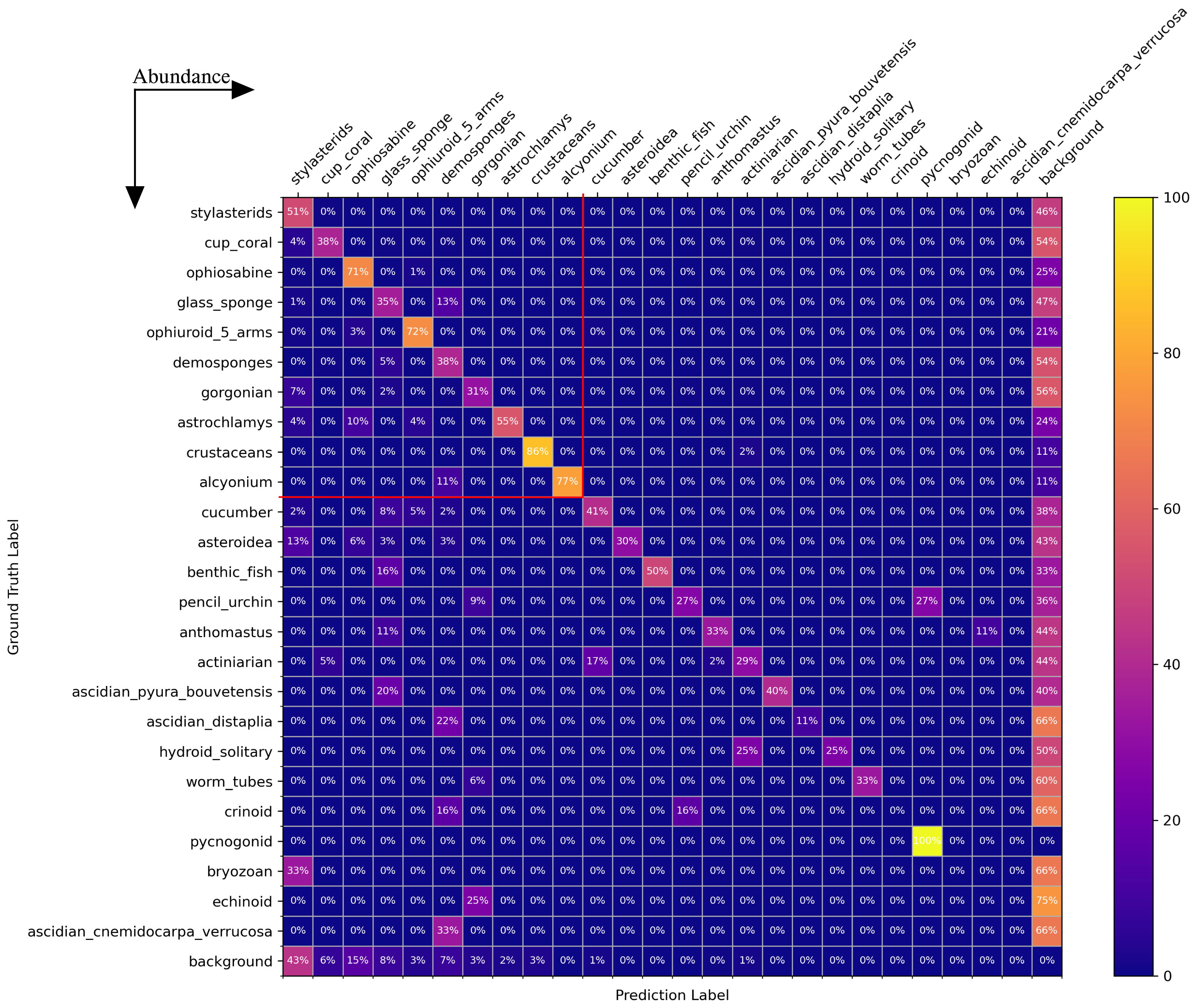}
\end{figure}

\newpage
\section{SAHI Postprocessing Limitations}\label{app:postprocessfail}

Large objects, split over a high number of patches, may fail to merge into a single coherent bounding box after Non-Maximum Merging via SAHI. See \ref{sec:discussion_Subsec:postprocessingFails} for further discussion. Confidence threshold $=$ 0.60.

\begin{figure}[H]
    \centering
    \caption{A single \texttt{ophiuroid\_5\_arms}, represented by two bounding boxes after postprocessing. Cropped and enlarged for clarity.}
    \includegraphics[width=0.75\linewidth]{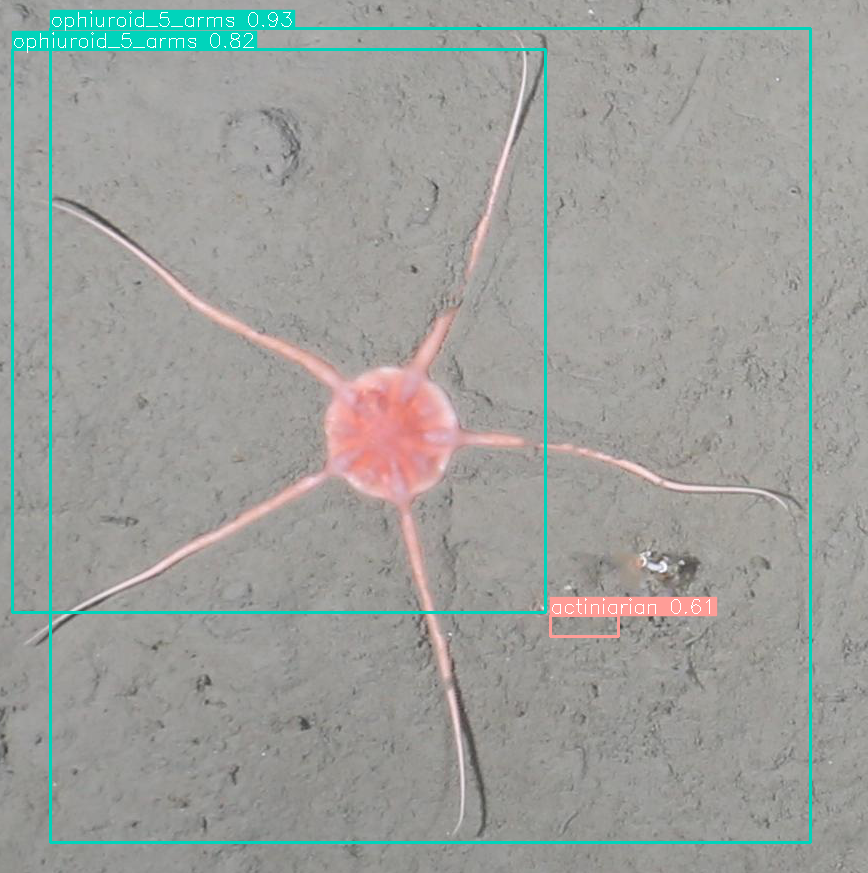}
\end{figure}

\end{document}